\documentclass{article}
\usepackage{spconf,amsmath,graphicx, amssymb,tabularx}
\usepackage[colorlinks=true]{hyperref}
\usepackage{subcaption}
\usepackage[font=small,skip=2pt]{caption}
\usepackage{multirow}
\usepackage{xcolor}
\usepackage{cite}
\usepackage{algpseudocode}
\usepackage{algorithm}
\usepackage{boxedminipage}
\usepackage[utf8]{inputenc}
\usepackage{enumitem}
 \usepackage{boxedminipage}
\title{Learning from past scans: Tomographic reconstruction\\ to detect new structures}
\name{Preeti Gopal$^{\star,\dagger}$ \qquad Sharat Chandran$^{\star}$ \qquad Imants
  Svalbe$^{\dagger}$ \qquad Ajit Rajwade$^{\star}$}
\address{
$^{\star}$Dept. of Computer Science \& Engg., Indian Institute of Technology Bombay \\
$^{\dagger}$School of Physics and Astronomy, Monash University\\
}
\begin{document}
%
\maketitle
\begin{abstract}
The need for tomographic reconstruction from sparse measurements
arises when the measurement process is potentially harmful, needs to
be rapid, or is uneconomical. In such cases, prior information from
previous longitudinal scans of the same or similar objects helps to reconstruct the current object whilst requiring significantly fewer `updating' measurements.
\textit{ However, a significant limitation of all prior-based methods is the possible dominance of the prior over the reconstruction of new localised information that has evolved within the test object.} In this paper, we improve the state of the art by (1) detecting potential regions where new changes may have occurred, and (2) effectively reconstructing both the old and new structures by computing regional weights that moderate the local influence of the priors. We have tested the efficacy of our method on synthetic as well as real volume data. The results demonstrate that using weighted priors significantly improves the overall quality of the reconstructed data whilst minimising their impact on regions that contain new information.
\end{abstract}
%
%
\section{Introduction}
\label{sec:intro}
\begin{figure}[h]
\centering
	\includegraphics[width=0.4\textwidth]{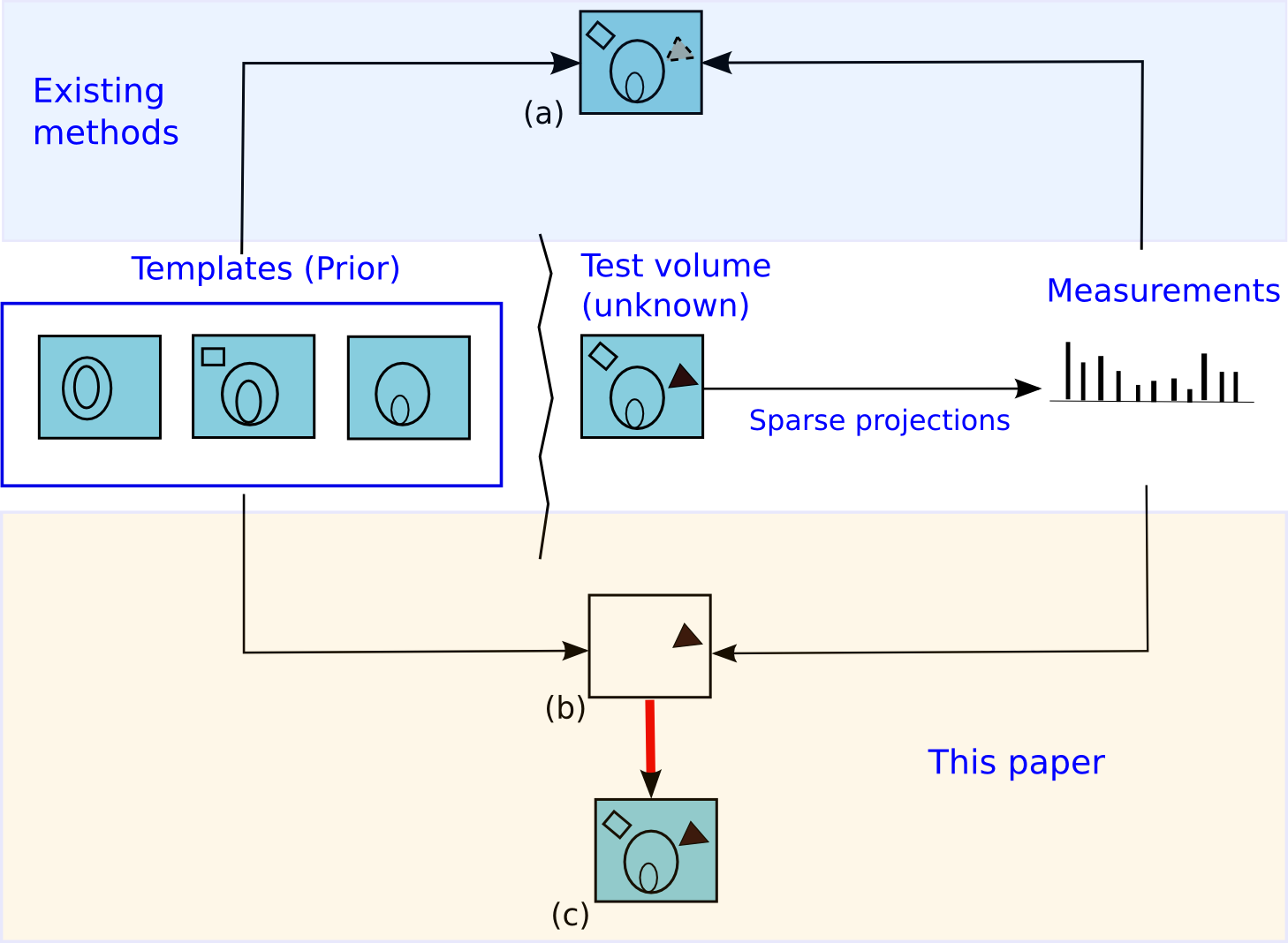}
        \caption{Overview of our method (a) Reconstruction with
          existing methods: new structures are not accurate
          (b) Spatially varying weights map (c) Our reconstruction.
          }
        \label{fig:prior_overview}
\end{figure}

Current research seeks to significantly reduce the number of tomographic measurements required to reconstruct an object with adequate fidelity. Sub-Nyquist sampling requires prior information or making some assumptions about the object shape. Compressive Sensing (CS)~\cite{Donoho} assumes sparsity in the computed image, often via information redundancy that is optimised through some mathematical transform. The measurement set can be reduced further by prior based techniques  \cite{PICCS,cardiacPICCS,lubner2011,pirple,mota2017}, that utilize previously scanned spatial data to reconstruct new volumes from sparse sets of additional measurements. A cost function used in these iterative reconstruction schemes penalizes any dissimilarity between the templates (prior) and the volume to be reconstructed (test). Another class of methods~\cite{van2009a,soma2011} impose information-theoretic similarity between the prior and the test volume. In all of these methods, it is critical and challenging to choose an optimal representative template.  Recent work~\cite{liu2016,Xu2012,my_dicta_paper} relaxed the above limitation by building dictionary based and eigenspace based priors. \cite{my_dicta_paper} showed that the eigenspace based prior is better able to accommodate the variation in the match of a test volume to a set of templates. The technique assumed the new test volume lies within the space spanned by the eigenvectors of multiple representative templates, effectively capturing the global properties of this set of templates. However, even with that approach, the global prior still imposes an inflexible constant weight when reconstructing the data. This results in inaccurate reconstruction for those critical regions where the test data may be marginally different from any of the prior data. In all prior based methods, the overall reconstruction improves, however, there is a likelihood of suppressing new features in the test object that may develop over time (refer Fig.~\ref{fig:prior_overview}). This scenario is particularly relevant in longitudinal studies, wherein the same object is scanned on multiple occasions to monitor for small changes that may occur over time.

\textbf{Contributions:} In this work, we use, in a novel way,
\emph{sparse} projections of the test volume to first detect the
regions in the test volume that may be different from all
priors. Those regions are the most likely ones to be of interest in
any longitudinal study.  We compute spatially varying weights to
temper the role played by the priors across different regions of the
object. With these weights, we blend the advantages of existing
reconstruction algorithms to create an optimal reconstruction.
Our reconstructions are validated on new, real, biological 3D
datasets, and existing medical datasets. 
\section{Weighted prior based reconstruction}

When an object is scanned multiple times, a set of high quality
reconstructions may be chosen as templates for the reconstruction of
future scan volumes, which, in turn, may be scanned using far fewer
measurements. The eigenspace $E_{\text{high}}$ of the $L$ prior
templates $Q_1,Q_2,...,Q_L$ is pre-computed. This is used as an orthogonal basis to represent the unknown test volume. While the prior
compensates very well for the new sparse measurements, it dominates
the regions with new changes masking the signal. Our weighted
prior based reconstruction overcomes (details appear
\href{https://www.dropbox.com/sh/mhc7grq9gcy2jws/AABNBpdswfLlz8UCNuxCuXKra?dl=0}{here}) this limitation by minimizing:
\begin{equation}
\begin{split}
   \setlength{\belowdisplayskip}{0pt} \setlength{\belowdisplayshortskip}{0pt}
\setlength{\abovedisplayskip}{-2pt} \setlength{\abovedisplayshortskip}{-2pt}
E(\boldsymbol{\theta},\boldsymbol{\alpha}) =& \lVert\boldsymbol{\Phi x- y}\rVert_2^2  + \lambda_1\lVert\boldsymbol{\theta}\rVert_1 \\&+\lambda_2\lVert\boldsymbol{W}(\boldsymbol{x} - (\boldsymbol{\mu} + \sum_{i}\boldsymbol{V_i}\alpha_i))\rVert_2^2
\end{split}
\label{eq:weighted_prior}
\end{equation}
Here, $\boldsymbol{x} = \boldsymbol{\Psi\theta}$ denotes the reconstructed volume, $\boldsymbol{y}$ its measured tomographic projections, $\boldsymbol{\theta}$ the sparse coefficients of $\boldsymbol{x}$, $\boldsymbol{\Psi}$ the
basis in which data is assumed to be sparse, $\boldsymbol{\Phi}$ a matrix modelling
the measurement geometry, $\boldsymbol{\mu}$ the mean of the templates,
$\boldsymbol{V_i}$ the $i^{th}$ principal component of the set of templates, $\boldsymbol{\alpha}$ the eigen coefficients, and $\lambda_1$, $\lambda_2$ are tunable weights given to the sparsity and prior terms respectively. 
 The unknowns $\boldsymbol{\theta}$ and $\boldsymbol{\alpha}$
are solved by alternate minimization. $\boldsymbol{\theta}$ is solved
for using the basis pursuit CS solver~\cite{l1ls}.

The key to our method is the discovery of the weights
$\boldsymbol{W}$, a matrix with  varying weights assigned to individual voxels of the prior. $\boldsymbol{W}$ is first
constructed using some preliminary reconstruction methods, following
which Eqn.~\ref{eq:weighted_prior} is used to get the final
reconstruction. In regions of change in test data,
we want lower weights for the prior when compared to regions that are
similar to the prior.  

\subsection{Computation of weights matrix $\boldsymbol{W}$}

The test volume is unknown to begin with. Hence, it is not possible to
decipher the precise regions in it that are new, and different from
all the templates. Schematic 1 describes the evolution of the
procedure used to detect the new regions in the unknown volume. We
start with $X^{\text{fdk}}$, the initial backprojection reconstruction
of the test volume using the Feldkamp-Davis-Kress (FDK)
algorithm~\cite{FDK} in an attempt to discover the difference between
the templates and the test volume.

However, the difference between $X^{\text{fdk}}$ and its projection
onto the eigenspace $E_{\text{high}}$ will detect the new regions
along with  many false positives (false new regions). This is because,
while $X^{\text{fdk}}$ has many artefacts arising from sparse
measurements, the eigenspace $E_{\text{high}}$ is built from high quality
templates. To discover unwanted artifacts of the imaging geometry, in a counter
intuitive way, we generate \emph{low quality} reconstruction of the templates.

\noindent \begin{boxedminipage}[t!]{0.5\textwidth} 
{\bf Schematic 1}: Motivation behind our algorithm. (The plus $\oplus$ and the
minus $\ominus$ operators are placeholders; precise details available
in Section~\ref{sec:thealgo}). 

{\small
 Let \textcolor{blue}{prior $Q:=$ old regions ($O$)}\\
 Let \textcolor{blue}{test volume $\boldsymbol{x}:=$ old regions ($O$) $\oplus$ new regions ($N$)}\\
    \begin{enumerate}[noitemsep]
\item  Compute pilot reconstruction of $\boldsymbol{x}$. Let this be called $X$.\\ \textcolor{blue}{$X = O \oplus N \oplus Ar(O) \oplus Ar(N)$}, where \\ \textcolor{blue}{$Ar(O)$} denote the reconstruction artefacts that depend on the old regions, the imaging geometry and the reconstruction method, and\\
 \textcolor{blue}{$Ar(N)$} denote the reconstruction artefacts that depend on the new regions, imaging geometry and the reconstruction method.
\item  Note that \textcolor{blue}{$Q \ominus X = N \oplus Ar(O) \oplus Ar(N)$} gives the new regions, but along with lots of artefacts due to the imaging geometry (sparse views). To eliminate these unwanted artefacts, compute \textcolor{blue}{$Y = Q \oplus Ar(O)$} by simulating projections from $Q$ using the same imaging geometry used to scan $\boldsymbol{x}$, and then reconstructing a lower quality prior volume $Y$. 
    \item Note that \textcolor{blue}{$Y \ominus X = N \oplus Ar(N)$} contains the artefacts due to the new regions only. These are different for different reconstruction methods. To eliminate these method dependent artefacts, compute $Y$ and $X$ using different reconstruction methods. Let these be denoted by $Y^j$ and $X^j$ respectively.
    \item Compute
\vspace{-0.2cm}
           \textcolor{blue}{\begin{equation*}
            Y^1 \ominus X^1 = N \oplus Ar^1(N)
           \end{equation*}
\vspace{-0.5cm}
           \begin{equation*}
            Y^2 \ominus X^2 = N \oplus Ar^2(N)
           \end{equation*}}
\vspace{-0.4cm}
   \item  New regions are obtained by computing
\vspace{-0.2cm}
          \textcolor{blue}{\begin{equation*}
          (Y^1 \ominus X^1)\cap(Y^2 \ominus X^2)= N
          \end{equation*}}   
\vspace{-0.5cm}
    \item Finally, assign space-varying weights $\boldsymbol{W}$ based on step $5$.
    \end{enumerate}
\label{algo:newRegionDetection}
}
\end{boxedminipage}

\subsection{Algorithm to find $\boldsymbol{W}$}
\label{sec:thealgo}
\begin{enumerate}[noitemsep,wide=0pt, leftmargin=\dimexpr\labelwidth + 2\labelsep\relax]

\item Perform a preliminary reconstruction $X^{\text{fdk}}$ of the
  test volume using FDK.

\item Compute low quality template volumes $Y^\text{fdk}$. 
In Schematic 1, for ease of exposition, we assumed a single
template. In the sequel, we assume $L$ templates from which we build
an eigenspace.

\vspace{-0.1cm}
  \begin{enumerate}[noitemsep,wide=0pt, leftmargin=\dimexpr\labelwidth
    + 2\labelsep\relax]

  \item Generate simulated measurements $\boldsymbol{y_{Q_i}}$ for
    every template $Q_i$, using the exact same projections views and
    imaging geometry with which the measurements $\boldsymbol{y}$ of the
    test volume were acquired, and 
  \item Perform $L$  preliminary FDK reconstructions of each of the
    $L$ templates from 
    $\boldsymbol{y_{Q_i}}$.  Let this be denoted by
    $\{Y^{\text{fdk}}_i\}_{i=1}^L$.
  \end{enumerate}
  
\item Build eigenspace $E_{\text{low}}$ from $\{Y^{\text{fdk}}_i\}_{i=1}^L$.  Let
  $P^{\text{fdk}}$ denote projection of $X^{\text{fdk}}$
  onto $E_{\text{low}}$. The difference between $P^{\text{fdk}}$ and
  $X^{\text{fdk}}$ will not contain false positives due
  to imaging geometry, but will have false positives due to artefacts
  that are specific to the reconstruction method used. To resolve
  this, perform steps $4$ and $5$.
\item Project with multiple methods.
  \begin{enumerate}[noitemsep,wide=0pt, leftmargin=\dimexpr\labelwidth + 2\labelsep\relax]

  \item Perform pilot reconstructions of the test using $M$ different
    reconstruction algorithms\footnote{CS~\cite{lasso}, ART~\cite{art},
      SART~\cite{sart} and SIRT~\cite{sirt}}. Let this set be denoted
    as $X \triangleq \{X^j\}_{j=1}^M$ where $j$ is an index for the
    reconstruction method, and $X^1 = X^{\text{fdk}}$ 

  \item From $\boldsymbol{y_{Q_i}}$, perform reconstructions of the
    template $Q_i$ using the $M$ different afore-mentioned algorithms,
    for each of the $L$ templates. Let this set be denoted by $Y \triangleq
    \{\{Y_{i}^j\}_{j=1}^M\}_{i=1}^L$ where $Y^{1}_i =
    Y^{\text{fdk}}_i$, $\forall i=1..L$.

  \item For each of the $M$ algorithms (indexed by $j$), build an eigenspace
    $E_\text{low}^j$ from $\{Y_1^j,Y_2^j, \ldots, Y_{L}^j\}$. 

  \item Next, for each $j$,  project $X^j$  onto $E_{\text{low}}^j$. Let
    this projection be denoted by $P^j$. To reiterate, this captures
    those parts of the test volume that lie in the subspace
    $E_{\text{low}}^j$ (i.e., are similar to the template
    reconstructions). The rest, new changes and their reconstruction
    method-dependent-artefacts are not captured by this projection and
    need to be eliminated.
  \end{enumerate}
\item To remove all reconstruction method dependent false positives,
  we compute $\min_{j}(|X^j(x,y,z) - P^j(x,y,z)|)$ .  

\item Finally, the weight to prior for each voxel coordinate $(x,y,z)$ is given by
  \vspace{-0.1cm}
  \begin{equation*} 
    \boldsymbol{W}(x,y,z) = (1+k(\min_{j}|X^j(x,y,z) - P^j(x,y,z)|))^{-1}
  \end{equation*}
\end{enumerate}
\vspace{-0.3cm} 
For each voxel $(x,y,z)$, the weight $\boldsymbol{W}(x,y,z)$ must be
low whenever the preliminary test reconstruction $X^j(x,y,z)$ is
different from its projection $P^j(x,y,z)$ onto the prior’s
eigenspace, for \emph{every} method $j \in \{1,...,M\}$. This is
because it is unlikely that every algorithm would produce a
significant artefact at a voxel. $k$ decides the sensitivity of the
weights to the difference $|X^j(x,y,z) - P^j(x,y,z)|$ and hence it
depends on the size of the new regions we want to detect.  
We found that our final reconstruction results obtained by
solving Eqn.~\ref{eq:weighted_prior} were robust over a wide~\footnote{Please refer to
  \href{https://www.dropbox.com/sh/mhc7grq9gcy2jws/AABNBpdswfLlz8UCNuxCuXKra?dl=0}{supplementary material: goo.gl/D4YjMQ}
  for details.}
range of $k$ values.

\begin{figure}
    \begin{subfigure}[b]{0.24\linewidth}
        \includegraphics[width=0.8\textwidth]{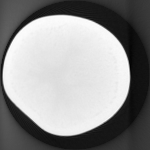}
\captionsetup{labelformat=empty}       
 \caption{}
    \end{subfigure}
    \begin{subfigure}[b]{0.24\linewidth}
        \includegraphics[width=0.8\textwidth]{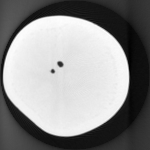}
\captionsetup{labelformat=empty}
        \caption{}
     \end{subfigure}
    \begin{subfigure}[b]{0.24\linewidth}
        \includegraphics[width=0.8\textwidth]{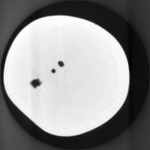}
\captionsetup{labelformat=empty}
        \caption{}
     \end{subfigure}
    \begin{subfigure}[b]{0.24\linewidth}
        \fcolorbox{yellow}{yellow}{\includegraphics[width=0.8\textwidth]{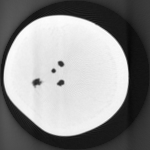}}
\captionsetup{labelformat=empty}
        \caption{}
     \end{subfigure}
      \caption{Potato 3D dataset: One slice each from the templates
        (the first three from left) and a slice from the test 
        volume (extreme right). Notice the appearance of the fourth
        hole in the test slice. }
\label{fig:templates_test_potato}
\end{figure}

\begin{figure}[h]
    \begin{subfigure}[b]{0.155\linewidth}
        \includegraphics[width=\textwidth]{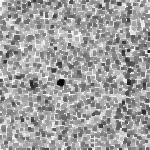}
        \caption{FDK}
    \end{subfigure}
    \begin{subfigure}[b]{0.155\linewidth}
        \includegraphics[width=\textwidth]{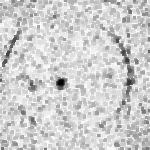}
        \caption{CS}
      \end{subfigure}
    \begin{subfigure}[b]{0.155\linewidth}
        \includegraphics[width=\textwidth]{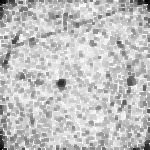}
        \caption{ART}
     \end{subfigure}
    \begin{subfigure}[b]{0.155\linewidth}
        \includegraphics[width=\textwidth]{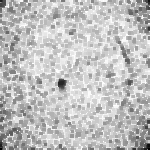}
        \caption{SART}
     \end{subfigure}
    \begin{subfigure}[b]{0.155\linewidth}
        \includegraphics[width=\textwidth]{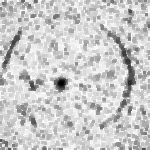}
        \caption{SIRT}
     \end{subfigure}
    \begin{subfigure}[b]{0.155\linewidth}
        \includegraphics[width=\textwidth]{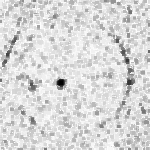}
 \captionsetup{singlelinecheck=false,format=hang}
        \caption{\mbox{Our method}}
     \end{subfigure}
      \caption{New regions (shown in lower intensities) detected by
        different reconstruction methods. These are different because
        the reconstruction artifacts of the new structures is different for every reconstruction method used. }
\label{fig:weights_map_2Dpotato}
\end{figure}
\begin{figure}
      \subcaptionbox{Test}{
        \fcolorbox{yellow}{yellow}{\includegraphics[width=0.175\linewidth]{images/potato/testIm.png}}}
      \subcaptionbox{FDK}{
        \includegraphics[width=0.18\linewidth]{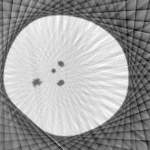}}
      \subcaptionbox{CS}{
        \includegraphics[width=0.18\linewidth]{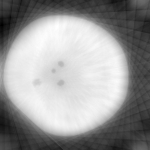}}
 \captionsetup{singlelinecheck=false,format=hang}
      \subcaptionbox{Plain prior}{
        \includegraphics[width=0.18\linewidth]{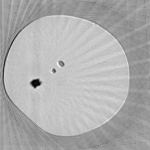}}
 \captionsetup{singlelinecheck=false,format=hang}
      \subcaptionbox{Our method}{
        \includegraphics[width=0.18\linewidth]{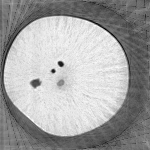}}
 \captionsetup{format=plain}
\caption{3D reconstruction of the potato with $5\%$ projection views--(b) has strong streak artefacts with unclear shadow of the potato, (c) largely blurred, (d) no new information detected (prior dominates) and (e) new information detected while simultaneously reducing streak artefacts.}  
\label{fig:potato_3D_results}
\end{figure}
\vspace{-0.4cm}
\section{Results}
Our algorithm is validated on new\footnote {These and our code will be made available to the community.}  scans of
biological specimens in a longitudinal setting. In all figures, `plain prior' refers to optimizing Eqn.~\ref{eq:weighted_prior} with $\boldsymbol{W}(x,y,z)=1$.

  \textbf{The first (Potato) dataset}
consisted of four scans of the humble potato, chosen for its
simplicity (Fig.~\ref{fig:templates_test_potato}). While the first
scan was taken of the undistorted potato, subsequent scans were taken
of the same specimen, each time after drilling a new hole halfway into
it.  Projections were obtained using circular cone beam geometry. The
specimen was kept in the same position throughout the acquisitions. In
case where this alignment is not present, all the template volumes
must be pre-aligned before computing the eigenspace. The test must be
registered to the templates after its preliminary pilot
reconstruction. Further, this alignment can be refined following every
update of the reconstructed test volume. The ground truth consists of
FDK reconstructions from the full set of acquired measurements from
900 projection views. The test volume was reconstructed using
measurements from 45 projection views, i.e, $5\%$ of the projection
views from which ground truth was reconstructed.  The selected 3D
ground truth of template volumes, test volume, as well as the 3D
reconstructions are shown \href{https://www.dropbox.com/sh/mhc7grq9gcy2jws/AABNBpdswfLlz8UCNuxCuXKra?dl=0}{here}. As seen
in Fig.~\ref{fig:potato_3D_results}, our method reconstructs new
structures while simultaneously reducing streak artefacts. The new
regions detected by different reconstruction methods are shown in
Fig.~\ref{fig:weights_map_2Dpotato}. The lower intensities denote new
regions which were assigned lower weights. This ensures that the new
regions are reconstructed using projection measurements alone.

\begin{figure}
    \begin{subfigure}[b]{0.18\linewidth}
        \includegraphics[width=0.8\textwidth]{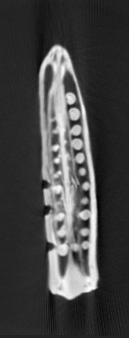}
\captionsetup{labelformat=empty}       
 \caption{}
    \end{subfigure}
    \begin{subfigure}[b]{0.18\linewidth}
        \includegraphics[width=0.8\textwidth]{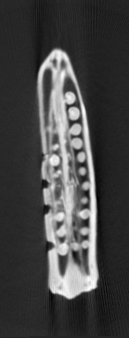}
\captionsetup{labelformat=empty}
        \caption{}
     \end{subfigure}
    \begin{subfigure}[b]{0.18\linewidth}
        \includegraphics[width=0.8\textwidth]{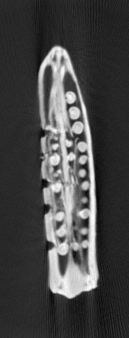}
\captionsetup{labelformat=empty}
        \caption{}
     \end{subfigure}
    \begin{subfigure}[b]{0.18\linewidth}
        \includegraphics[width=0.8\textwidth]{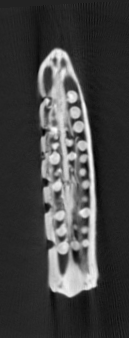}
\captionsetup{labelformat=empty}
        \caption{}
     \end{subfigure}
    \begin{subfigure}[b]{0.176\linewidth}
        \fcolorbox{yellow}{yellow}{\includegraphics[width=0.8\textwidth]{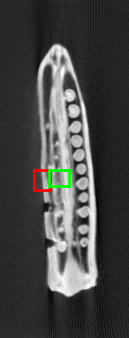}}
\captionsetup{labelformat=empty}
        \caption{}
     \end{subfigure}
     \caption{Okra 3D dataset: One slice each from the templates (the
       first four from the left), and one from the test volume
       (extreme right). In the regions marked in red and green, while
       all slices have deformities, the test
       has none.}
\label{fig:templates_test_okra}
\end{figure}

\begin{figure}[h]
 \captionsetup{singlelinecheck=false,format=hang}
      \subcaptionbox{Test}{
        \fcolorbox{yellow}{yellow}{\includegraphics[width=0.176\linewidth]{images/okra/testCropped.png}}}
      \subcaptionbox{FDK}{
        \includegraphics[width=0.18\linewidth]{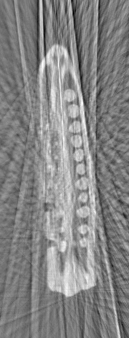}}
      \subcaptionbox{CS}{
        \includegraphics[width=0.18\linewidth]{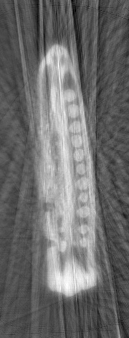}}
      \subcaptionbox{Plain prior}{
        \includegraphics[width=0.18\linewidth]{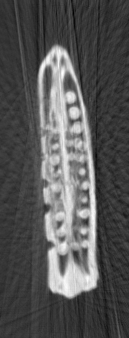}}
      \subcaptionbox{Our method}{
        \includegraphics[width=0.18\linewidth]{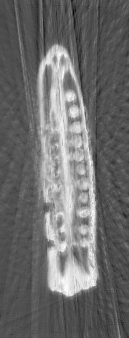}}
 \captionsetup{format=plain}
\caption{3D reconstruction of the okra from $10\%$ projection
  views (b) has strong streak artefacts, (c) blurred, (d) no new
  information detected (prior dominates -- the deformity from the prior
  shows up as a false positive) and (e) new information detected (no deformities
  corresponding to red and green regions) while simultaneously
  reducing streak artefacts.} 
\label{fig:okra_3D_results}
\end{figure}

In order to test on data with
intricate structures, a \textbf{second (Okra) dataset} consisting of
five scans of an okra specimen was acquired
(Fig.~\ref{fig:templates_test_okra}). The projections were obtained by
circular cone beam projection. Prior to the first scan, two holes were
drilled on the surface of the specimen. This was followed by four
scans, each after introducing one new cut. The ground truth consists
of FDK reconstructed volumes from the the full set of 450 view
projections. The test volume was reconstructed from a partial set of 45
projections, i.e, $10\%$ of the projection views from which ground
truth was reconstructed. The selected 3D ground truth of template
volumes, the test volume as well as the 3D reconstructions can be seen
\href{https://www.dropbox.com/sh/mhc7grq9gcy2jws/AABNBpdswfLlz8UCNuxCuXKra?dl=0}{here}. One of the slices of the
reconstructed volumes is shown in Fig.~\ref{fig:okra_3D_results}. The
red and green 3D RoI in the video and images show the regions where
new changes are present.

Based on the potato and the okra experiments, we see that our method
is able to \emph{discover both} the presence of a new structure, as
  well as the absence of a structure.


\begin{figure}[b]
    \begin{subfigure}[b]{0.155\linewidth}
        \includegraphics[width=\textwidth]{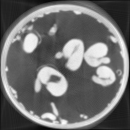}
\captionsetup{labelformat=empty}
        \caption{}
    \end{subfigure}
    \begin{subfigure}[b]{0.155\linewidth}
        \includegraphics[width=\textwidth]{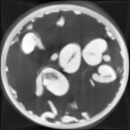}
\captionsetup{labelformat=empty}
        \caption{}
     \end{subfigure}
    \begin{subfigure}[b]{0.155\linewidth}
        \includegraphics[width=\textwidth]{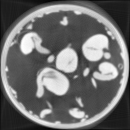}
\captionsetup{labelformat=empty}
        \caption{}
     \end{subfigure}
    \begin{subfigure}[b]{0.155\linewidth}
        \includegraphics[width=\textwidth]{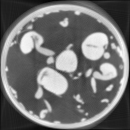}
\captionsetup{labelformat=empty}
        \caption{}
     \end{subfigure}
    \begin{subfigure}[b]{0.155\linewidth}
        \includegraphics[width=\textwidth]{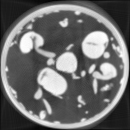}
\captionsetup{labelformat=empty}
        \caption{}
     \end{subfigure}
    \begin{subfigure}[b]{0.15\linewidth}
        \fcolorbox{yellow}{yellow}{\includegraphics[width=\textwidth]{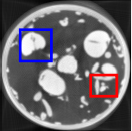}}
\captionsetup{labelformat=empty}
        \caption{}
     \end{subfigure}
      \caption{Sprouts 3D dataset: One slice each from the templates
        (the first five from left) and a slice from the test (extreme
        right).}
\label{fig:templates_test_sprouts}
\addtolength{\textfloatsep}{-0.8cm}
\end{figure}

\textbf{The third (Sprouts) dataset} consists of six scans of a
sprouts specimen imaged at its various stages of growth
(Fig.~\ref{fig:templates_test_sprouts}). In contrast to the scientific
experiment performed for the case of the okra and the potato where we
introduced man-made defects, the changes here are purely the work of
nature.

\begin{figure}[h!]
      \subcaptionbox{Test}{
        \fcolorbox{yellow}{yellow}{\includegraphics[width=0.176\linewidth]{images/sprouts/testIm.png}}}
      \subcaptionbox{FDK}{
        \includegraphics[width=0.18\linewidth]{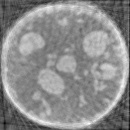}}
      \subcaptionbox{CS}{
        \includegraphics[width=0.18\linewidth]{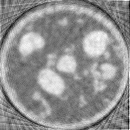}}
 \captionsetup{singlelinecheck=false,format=hang}
      \subcaptionbox{Plain prior}{
        \includegraphics[width=0.18\linewidth]{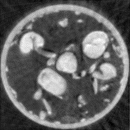}}
      \subcaptionbox{Our method}{
        \includegraphics[width=0.18\linewidth]{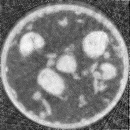}}
 \captionsetup{format=plain}
 \caption{3D reconstruction of sprouts from $2.5\%$ projection views
   (b, c) have poor details (d) no new information detected (the
   prior dominates as can be seen in the blue and red regions) and
   (e) new information detected in the regions of interest.} 
\label{fig:sprouts_3D_results}
\end{figure}
\addtolength{\textfloatsep}{-0.2cm}

The ground truth consists of FDK reconstructed volumes from a set of
1800 view projections. The test volume was reconstructed from partial
set of 45 projections, i.e, $2.5\%$ of the projection views from which
ground truth was reconstructed. The selected 3D ground truth of
template volumes, test volume, as well as the 3D reconstructions are
shown \href{https://www.dropbox.com/sh/mhc7grq9gcy2jws/AABNBpdswfLlz8UCNuxCuXKra?dl=0}{here}. One of the slices of the
reconstructed volumes is shown in
Fig.~\ref{fig:sprouts_3D_results}. For the sake of exposition, the red
and blue regions of interest (RoI) have been culled out from 7
consecutive slices in the 3D volume to indicate new structures; other
changes can be viewed in the video.

\setlength{\tabcolsep}{5pt}


\begin{table}[h]
\caption{Quantitative SSIM values (a value of 1 is the best
  possible). For targeted and clear quantitative comparison, the
intricate regions of interests in the okra and the sprouts have been
used.} 
\begin{tabularx}{0.94\linewidth}{|c|c|c|c|c|}
\hline
\textbf{} & \textbf{FDK} & \textbf{CS} & \textbf{Plain prior} & \textbf{Our method} \\ \hline
\textbf{Potato} & 0.744 & 0.817 & 0.856 & {\color[HTML]{3531FF} \textbf{0.857}} \\ \hline
\textbf{Okra} & 0.737 & 0.836 & 0.858 & {\color[HTML]{3531FF} \textbf{0.883}} \\ \hline
\textbf{Sprouts} & 0.852 & 0.843 & 0.834 & {\color[HTML]{3531FF} \textbf{0.881}} \\ \hline
\end{tabularx}
\label{table:all_ssim}
\end{table}


Table~\ref{table:all_ssim} shows the improvement in the Structure
Similarity Index (SSIM) of the reconstructed new regions as compared
to other methods. 

\vspace{-0.6cm}
\section{Conclusions}
In a longitudinal study, the reconstruction of localized new
information in the data should not be affected by priors used, given
that the new measurements are taken with substantially fewer views
(approximately 2.5\%--10\% data). In this work, we have improved the
state of the art with a
weighted prior-based reconstruction that detects these regions of
change and assigns low prior weights wherever necessary. The
probability of presence of a `new region' is enhanced considerably by
a novel combination of different reconstruction techniques. Our method
is general as shown in Schematic~1, but has been demonstrated on new,
real, biological 3D datasets for longitudinal studies. The method is also largely robust to the number of templates used. We urge the
reader to see the videos of reconstructed volumes in the
\href{https://www.dropbox.com/sh/mhc7grq9gcy2jws/AABNBpdswfLlz8UCNuxCuXKra?dl=0}{supplementary material}~\cite{supp}.
In a medical setting, we note that the detection of the new regions
will enable irradiation of the patient only in the
region of interest by keeping specific detector bins
active, thereby reducing the radiation exposure.  
\newpage
\bibliographystyle{IEEEbib}
{
\bibliography{reconstruction_ref.bib}}

\begin{thebibliography}{10}

\bibitem{Donoho}
D.L. Donoho,
\newblock ``Compressed sensing,''
\newblock {\em IEEE Transactions on Information Theory}, vol. 52, no. 4, pp.
  1289--1306, April 2006.

\bibitem{PICCS}
Guang-Hong Chen, Jie Tang, and Shuai Leng,
\newblock ``Prior image constrained compressed sensing {(PICCS)}: A method to
  accurately reconstruct dynamic {CT} images from highly undersampled
  projection data sets,''
\newblock {\em Medical Physics}, vol. 35, no. 2, pp. 660--663, 2008.

\bibitem{cardiacPICCS}
G.~Chen, P.~Theriault-Lauzier, J.~Tang, B.~Nett, S.~Leng, J.~Zambelli, Z.~Qi,
  N.~Bevins, A.~Raval, S.~Reeder, and H.~Rowley,
\newblock ``Time-resolved interventional cardiac {C}-arm cone-beam {CT}: An
  application of the {PICCS} algorithm,''
\newblock {\em IEEE Transactions on Medical Imaging}, vol. 31, no. 4, pp.
  907--923, April 2012.

\bibitem{lubner2011}
Meghan~G. Lubner, Perry~J. Pickhardt, Jie Tang, and Guang-Hong Chen,
\newblock ``Reduced image noise at low-dose multidetector {CT} of the abdomen
  with prior image constrained compressed sensing algorithm,''
\newblock {\em Radiology}, vol. 260, pp. 248--256, July 2011.

\bibitem{pirple}
J~Webster Stayman, Hao Dang, Yifu Ding, and Jeffrey~H Siewerdsen,
\newblock ``{PIRPLE}: A penalized-likelihood framework for incorporation of
  prior images in {CT} reconstruction,''
\newblock {\em Physics in medicine and biology}, vol. 58, no. 21, Nov 2013.

\bibitem{mota2017}
J.~F.~C. Mota, N.~Deligiannis, and M.~R.~D. Rodrigues,
\newblock ``Compressed sensing with prior information: Strategies, geometry,
  and bounds,''
\newblock {\em IEEE Transactions on Information Theory}, vol. 63, no. 7, pp.
  4472--4496, July 2017.

\bibitem{van2009a}
Dominique~Van de~Sompel and Michael Brady,
\newblock ``Robust incorporation of anatomical priors into limited view
  tomography using multiple cluster modelling of the joint histogram,''
\newblock {\em 2009 IEEE International Symposium on Biomedical Imaging: From
  Nano to Macro}, pp. 1279--1282, 2009.

\bibitem{soma2011}
S.~Somayajula, C.~Panagiotou, A.~Rangarajan, Q.~Li, S.~R. Arridge, and R.~M.
  Leahy,
\newblock ``{PET} image reconstruction using information theoretic anatomical
  priors,''
\newblock {\em IEEE Transactions on Medical Imaging}, vol. 30, no. 3, pp.
  537--549, March 2011.

\bibitem{liu2016}
J.~Liu, Y.~Hu, J.~Yang, Y.~Chen, H.~Shu, L.~Luo, Q.~Feng, Z.~Gui, and
  G.~Coatrieux,
\newblock ``3{D} feature constrained reconstruction for low dose {CT}
  imaging,''
\newblock {\em IEEE Transactions on Circuits and Systems for Video Technology},
  vol. PP, no. 99, pp. 1--1, 2016.

\bibitem{Xu2012}
Q.~Xu, H.~Yu, X.~Mou, L.~Zhang, J.~Hsieh, and G.~Wang,
\newblock ``Low-dose {X}-ray {CT} reconstruction via dictionary learning,''
\newblock {\em IEEE Transactions on Medical Imaging}, vol. 31, no. 9, pp.
  1682--1697, Sept 2012.

\bibitem{my_dicta_paper}
P.~Gopal, R.~Chaudhry, S.~Chandran, I.~Svalbe, and A.~Rajwade,
\newblock ``Tomographic reconstruction using global statistical priors,''
\newblock in {\em Digital Image Computing: Techniques and Applications
  (DICTA)}, Sydney, Australia, Nov. 2017.

\bibitem{l1ls}
K.~Koh, S.-J. Kim, and S.~Boyd,
\newblock ``l1-ls: Simple matlab solver for l1-regularized least squares
  problems,'' \url{https://stanford.edu/~boyd/l1_ls/}, last viewed--July, 2016.

\bibitem{FDK}
Lee Feldkamp, L.~C. Davis, and James Kress,
\newblock ``Practical cone-beam algorithm,''
\newblock {\em J. Opt. Soc. Am}, vol. 1, pp. 612--619, 01 1984.

\bibitem{lasso}
Robert Tibshirani,
\newblock ``Regression shrinkage and selection via the lasso,''
\newblock {\em Journal of the Royal Statistical Society. Series B
  (Methodological)}, vol. 58, no. 1, pp. 267--288, 1996.

\bibitem{art}
Richard Gordon, Robert Bender, and Gabor~T. Herman,
\newblock ``Algebraic reconstruction techniques {(ART)} for three-dimensional
  electron microscopy and {X}-ray photography,''
\newblock {\em Theoretical Biology}, vol. 29, no. 3, pp. 471--481, Dec 1970.

\bibitem{sart}
A.H. Andersen and A.C. Kak,
\newblock ``Simultaneous algebraic reconstruction technique ({SART}): A
  superior implementation of the {ART} algorithm,''
\newblock {\em Ultrasonic Imaging}, vol. 6, no. 1, pp. 81 -- 94, 1984.

\bibitem{sirt}
Peter Gilbert,
\newblock ``Iterative methods for the three-dimensional reconstruction of an
  object from projections,''
\newblock {\em Journal of Theoretical Biology}, vol. 36, no. 1, pp. 105 -- 117,
  1972.

\bibitem{supp}
``{Supplementary material},''
  \url{https://www.dropbox.com/sh/mhc7grq9gcy2jws/AABNBpdswfLlz8UCNuxCuXKra?dl=0},
  last viewed--Oct, 2018.

\end{thebibliography}

\end{document}